\documentclass[10pt,twocolumn,twoside]{IEEEtran}

\usepackage{times}
\usepackage{epsfig}
\usepackage{graphicx}
\usepackage{amsmath}
\usepackage{amssymb}
\usepackage{psfrag}
\usepackage[caption=false]{subfig}

\usepackage{enumitem}
\usepackage[crop=pdfcrop,process=auto,cleanup={.tex, .dvi, .ps, .pdf, .log, .bbl, .spl, .aux},bitmap=lossless]{pstool}
\usepackage{url}
\usepackage[numbers]{natbib}
\usepackage{xfrac}
\usepackage{multirow}
\usepackage{siunitx}
\usepackage[super]{nth}
\usepackage[noadjust]{cite}

\renewcommand{\cite}{\citep}

\def\hyph{-\penalty0\hskip0pt\relax}

\begin{document}


\title{An Empirical Study into Annotator Agreement, Ground Truth Estimation, and Algorithm Evaluation}

\author{Thomas~A.~Lampert$^*$,
        Andr\'{e}~Stumpf,
        and~Pierre~Gan\c{c}arski
\thanks{Manuscript received September 28, 2014; revised October 26, 2015 and March 15, 2016; accepted March 17, 2016. This work was supported by the French Research Agency through the COCLICO Project: ANR Mod\`{e}les Num\'e{e}riques Program under Grant ANR-12-MN-001-COCLICO 2012--2016.}
\thanks{T.\ A.\ Lampert and P.\ Gan\c{c}arski are with the ICube Laboratory, University of Strasbourg, France (e-mail: tlampert@unistra.fr).}
\thanks{A.\ Stumpf is with the Laboratoire Image, Ville, et Environment (LIVE), University of Strasbourg, France.}
\thanks{\copyright{}2016 IEEE. Personal use of this material is permitted. Permission from IEEE must be obtained for all other uses, in any current or future media, including reprinting/republishing this material for advertising or promotional purposes, creating new collective works, for resale or redistribution to servers or lists, or reuse of any copyrighted component of this work in other works.}
}

\markboth{IEEE Transactions on Image Processing,~Vol.~25, No.~6, March~2016}%
{Lampert \MakeLowercase{\textit{et al.}}: An Empirical Study into Annotator Agreement, Ground Truth Estimation, and Algorithm Evaluation}


\IEEEtitleabstractindextext{%
\begin{abstract}
Although agreement between annotators who mark feature locations within images has been studied in the past from a statistical viewpoint, little work has attempted to quantify the extent to which this phenomenon affects the evaluation of foreground-background segmentation algorithms. Many researchers utilise ground truth in experimentation and more often than not this ground truth is derived from one annotator's opinion. How does the difference in opinion affect an algorithm's evaluation? A methodology is applied to four image processing problems to quantify the inter-annotator variance and to offer insight into the mechanisms behind agreement and the use of ground truth. It is found that when detecting linear structures annotator agreement is very low. The agreement in a structure's position can be partially explained through basic image properties. Automatic segmentation algorithms are compared to annotator agreement and it is found that there is a clear relation between the two. Several ground truth estimation methods are used to infer a number of algorithm performances. It is found that: the rank of a detector is highly dependent upon the method used to form the ground truth; and that although STAPLE and LSML appear to represent the mean of the performance measured using individual annotations, when there are few annotations, or there is a large variance in them, these estimates tend to degrade. Furthermore, one of the most commonly adopted combination methods---consensus voting---accentuates more obvious features, resulting in an overestimation of performance. It is concluded that in some datasets it is not possible to confidently infer an algorithm ranking when evaluating upon one ground truth.
\end{abstract}

\begin{IEEEkeywords}
Evaluation, ranking, performance, feature detection, agreement, annotation, ground truth, gold-standard ground truth, expert agreement, receiver operating characteristic analysis, precision, recall.
\end{IEEEkeywords}}

\maketitle

\IEEEdisplaynontitleabstractindextext

\ifCLASSOPTIONpeerreview
\begin{center} \bfseries EDICS Category: MLR-PATT \end{center}
\fi
%
\IEEEpeerreviewmaketitle


\section{Introduction}


The evaluation of computer vision algorithms often requires ground truth (GT) data. The difficulty presented by this is that a gold-standard GT can be costly to obtain (if at all possible). For example, determining gold-standard GT in remote sensing experiments would typically require field surveys over large and sometimes remote areas and for medical scans difficulties arise since it would require invasive surgery. It is therefore commonly assumed that the opinion of one (or more) annotator(s) approximates this gold-standard GT. Nevertheless, annotators rarely agree completely when giving their opinion and this disagreement can be characterised as bias---the tendency of an annotator to prefer one decision over another---and variance---the natural variation that one annotator will have to the next (or themselves at a later date) \cite{Warfield08}. This poses a problem when evaluating computer vision algorithms: how does the difference in annotator opinion affect an algorithm's evaluation?

This work intends to quantify the effects of GT variability on the design, training, and evaluation of segmentation algorithms. To this end, supervised and unsupervised algorithms are evaluated in four case studies, all of which embody typical computer vision problems: the segmentation of natural images (referred to as the Segmentation case study), the identification of fissures in aerial imagery (referred to as the Fissure case study), the identification of landslides in satellite imagery (referred to as the Landslide case study), and the identification of blood vessels in medical imagery (referred to as the Blood Vessel case study). The true GT of these data sets (the gold-standard GT) cannot be deduced from the imagery alone and annotations by human experts are used as the best available approximation. This limitation is typical in many computer vision applications such as medical imaging, remote sensing, and natural scene analysis. Furthermore, there exist many objects in these datasets that can cause false-positive and false-negative errors, making them ideal to study annotator and detector agreements.

Several previous studies have developed statistical methods for estimating the gold-standard GT from a number of annotations \cite{Warfield08, Biancardi09, Burl94, Kauppi09, Langerak10, Li11, Smyth94, Warfield04}. Although some public datasets offer segmentations obtained from different annotators \cite{Arbelaez11, Armato11, Hoover00} these methods are rarely employed in real-world algorithm evaluation, where experimentation is typically limited to one annotation. Consequently, little is known about how different GTs and estimated gold standards affect the performance comparison of different algorithms. 

Through performance evaluation, GT data often influences an algorithm's design, the choice of an algorithm's parameter values, and also influences the structure of the training data itself. It is therefore important to quantify the effect that different GTs have on reporting an algorithm's performance. Relying on one annotator's opinion allows an algorithm to learn the annotator's bias, and does not necessarily result in a model that is effective at locating the true target. This problem can be circumvented when the images are captured in tightly controlled conditions or are synthetically generated \cite{Lampert11} because the gold-standard GT is trivial to calculate. In remote sensing and medical imaging problems, and those concerning natural images, this is not the case.

The following assumptions regarding the problem's characteristics are implicitly made within this study. In computer vision problems true positive locations tend to be spatially correlated (segments tend not to be lone pixels, but a number of connected pixels) and correlated with some image properties. It is assumed that the annotators are not malicious in producing their annotation, are not producing annotations at random, and are not simply following low-level cues in the image, but are instead able to draw upon some higher-level knowledge. This allows them to distinguish between segments that belong to the negative class, but share the same low-level image properties as those segments that constitute the positive class.

Therefore the objectives of this study are to: 
\begin{itemize}[noitemsep,nolistsep]
  \item empirically demonstrate any bias that results from evaluating an algorithm with a single annotation;
  \item quantify the effect that different GTs may have on the evaluation of multiple algorithms;
  \item and provide a general comparison between algorithms designed to infer the gold-standard GT.
\end{itemize}

The following section reviews relevant work from the literature. Section \ref{sec:methodology} prescribes the experimental methodology, the analysed datasets and the results are described in Section \ref{sec:experiments}, and a discussion of these results is presented in Section \ref{sec:discussion}. Finally, Section \ref{sec:conclusions} presents the study's conclusions.


\section{Related Work}
\label{sec:related}

In a classic study \citet{Smyth94} analyse the uncertainty of an annotator's judgement in marking volcanoes in synthetic aperture radar images of Venus. The authors assume a stochastic labelling process, to account for intra-annotator variability, and outline the probabilistic free-response ROC analysis that integrates the uncertainty of an annotator's judgement directly into the performance measure.

More recently a number of methods for combining multiple image annotations are proposed. These include work from the medical domain in which practitioners manually segment anatomical scans. The annotations are subsequently warped to match novel scans in order to estimate their segmentations. \citet{Kauppi09} take GTs as the intersection (consensus), fixed size neighbourhoods of the points marked by each annotator, and a combination of the two. The authors conclude that the intersection method is preferential as it results in the highest detector performance. Numerous weighted extensions to the voting framework have been proposed based upon global \cite{Sabuncu10}, local \cite{Artaechevarria09, Isgum09, Sabuncu10}, semi-local \cite{Sabuncu10, Wang13}, and non-local \cite{Coupe11} information.

Probably the most popular gold-standard GT estimation method originating from the medical domain is proposed by \citet{Warfield04}, named simultaneous truth and performance level estimation (STAPLE) in which annotator performance (measured as sensitivity and specificity) and the gold-standard GT are simultaneously estimated within a maximum\hyph{}likelihood setting, the optimisation being solved using expectation\hyph{}maximisation  (a variant for handling continuous labels has been proposed by \citet{Warfield08} and \citet{Xing11}). The same authors also propose an approach in which the bias and variance of each annotator is estimated instead of their sensitivity and specificity \cite{Warfield08} and another variant that accounts for instabilities in the annotator performance measures \cite{Commowick10}. Much subsequent work has concentrated on the STAPLE algorithm: removing its assumption that annotator performances are constant throughout the data \cite{Asman11, Asman12a, Commowick12}, and COLLATE \cite{Asman11b}, which accounts for spatial variability in task difficulty. \citet{Landman10} point out that in research and clinical environments it is not often possible to obtain multiple annotations for the whole dataset. Extensions to handle multiple partial, but overlapping, annotations have therefore been proposed \cite{Commowick10, Landman10, Landman13}.
 
\citet{Kamarainen12} propose a simpler alternative to STAPLE by maximising the mutual agreement of annotator ratings. This approach avoids the use of priors, and does not introduce segments that did not appear in the original annotations. \citet{Langerak10} argue, however, that STAPLE fails when annotator uncertainty varies considerably due to the fact that the STAPLE algorithm combines all of the annotators' labellings. Instead they propose the selective and iterative method for performance level estimation (SIMPLE) in which only labels that are deemed reliable are taken into account. \citet{Li11} propose a probabilistic approach that uses level sets in which the likelihood function is inspired by the STAPLE algorithm (LSML). To overcome the susceptibility of the STAPLE algorithm to strongly diverging annotations they accept that the contribution of an annotator's judgement should be dependent upon their performance, but differently to STAPLE the energy function is constrained by a shape prior that is dependent upon the amount of detail in the annotator's marking, forming the LSMLP algorithm. \citet{Biancardi09} state that the STAPLE algorithm (even with the Markov random field extension) and simple voting strategies assume that the pixels are spatially independent. A novel voting procedure is introduced to overcome this. It is preceded by a distance transformation that attributes positive values to the inside of the GT segmentation's boundary, which increase towards its centre, and decreases negatively outside the segment border; thus the truth estimate from self distances (TESD) algorithm is introduced \cite{Biancardi09}.

A new direction that has recently gained interest is to combine the information derived from the manual annotations with that derived from the image to imply the location of features-of-interest. \citet{Yang11} follow this path and propose a method that incorporates the warping error to preserve topological disagreements between the estimated gold-standard GT and the annotations. A number of extensions to the STAPLE algorithm have also been proposed \cite{Asman12, Asman13, Liu13} which incorporate the image's intensity values, as well as the performance of multiple experts, to transfer the labelling of one image onto that of another. Moreover, \citet{Landman12} propose to combine a locally weighted voting strategy with information derived form the image's intensity.

The widely used Berkeley segmentation dataset contains five-hundred images, each having five GTs. The authors include the level of annotator agreement within their evaluations \cite{Arbelaez11}, which provides a valuable reference when interpreting the results. Using the earlier Berkeley 300 database, \citet{Martin01} present a statistical analysis of the variation observed within the annotations \cite{Martin01}. They notice that independent annotators tend to include the same pixel in the same region, but also that the number of segments in the same image can vary by a factor of ten. The impact of GTs from different annotators on the ranking of segmentation algorithms has not yet been investigated.

\section{Methodology}
\label{sec:methodology}

To recapitulate, this work aims to demonstrate the effects of GT variability on the design, training, and evaluation of segmentation algorithms by studying their performance measured using single annotations, comparing multiple algorithms using different GTs, and comparing gold-standard GT inference algorithms. To achieve these aims, the methodological evaluation will be centred around three aspects: annotator agreement; the relation between annotator agreement and detector performance; and ground truths and reported detector performance. Scripts to recreate the results presented henceforth are available on-line\footnote{\url{https://sites.google.com/site/tomalampert/code}}.

\subsection{Data}

The data used in each of the case studies can be modelled as an image, $I : \{0,1,\dots,X-1\}\times\{0,1,\dots,Y-1\} \mapsto \mathbb{R}$ where $X$ is the image's width and $Y$ its height.

For each study $N$ annotators have provided manual markings containing the locations of the foreground target specific to each study. All case studies are binary detection problems and each annotation has the value one where the annotator perceived the feature-of-interest to exist and zero otherwise. The result of this process is are $N$ binary maps describing the location of the features-of-interest according to each annotator. As such, each annotator's output is modelled as a function $M_n : \{0,1,\dots,X-1\}\times\{0,1,\dots,Y-1\} \mapsto \{0,1\}$, where $0$ and $1$ represent the absence and presence of the object respectively and $n = 1,\dots, N$.


\subsection{Annotator Agreement}
\label{sec:annotator_agreement}

The first stage of analysis tests the level of agreement between the annotators in each case study, and exposes the image properties that promote this agreement.

\citet{Smyth96} presents a method for calculating the lower bound on error in a set of annotations relative to the (unknown) gold-standard ground-truth. This bound is defined to be
\begin{equation}
\bar{e} \ge \frac{1}{XYN} \sum^{Y-1}_{y=0} \sum^{X-1}_{x=0} \min \left \{N-A(x,y), A(x,y) \right \}
\label{eq:smyth}
\end{equation}
where $A(x,y)$ is the number of annotators that labelled pixel $(x,y)$ as containing the feature-of-interest, Equation \eqref{eq:agreement}. The minimum of Equation \eqref{eq:smyth} is reached when all annotators agree and the maximum (\num{0.5}) when the decisions are evenly split. It is therefore closely related to the entropy of the annotators' decisions. The maximum value for an acceptable level of experimental data quality suggested by the author is \SI{10}{\percent}.

Also to this end, the per-pixel annotator agreement is calculated, which is simply the number of annotators that have marked each pixel, such that
\begin{equation}
A(x,y) = \sum_{n = 1}^{N}{M_n(x,y)}.
\label{eq:agreement}
\end{equation}
The ratio of the number of pixels that are have a minimum level of agreement to the number of pixels that belong to annotated regions can therefore be calculated as follows
\begin{equation}
\hat{A}(n) = \frac{1}{|C|}  \sum_{x = 0}^{X-1} \sum_{y = 0}^{Y-1} {\chi_{B}(x,y) }
\label{eq:agreement_marker}
\end{equation}
where ${B} = \{(x,y) \mid A(x,y) \ge n\}$, $\chi_B$ is the indicator function, $C = \{(x,y) \mid A(x,y) > 0\}$, and $1 \le n \le N$ is the range of values for the minimum level of agreement. 

These functions allow for the testing of correlations between annotator agreement and different image properties---a means to uncover at least part of the reason behind the variance of agreement. Each dataset presents different features, but where applicable the following will be tested: intensity, contrast, and each of the colour channels. The Pearson's $r$ correlation coefficient will be used and, since the sample size for the analysis is extremely large, it will be tested for significance to \SI{99}{\percent} confidence. In the case that the image is colour, intensity is calculated such that $I(x,y) = 0.2989 \cdot R(x,y) + 0.5870 \cdot G(x,y) + 0.1140 \cdot B(x,y)$. Image contrast in a colour image is calculated using the Michelson contrast measure within a $3 \times 3$ local neighbourhood such that
\begin{equation}
c(x,y) = \frac{\max_{(i,j) \in W_{xy}} L(i,j) - \min_{(i,j) \in W_{xy}} L(i,j)}{\max_{(i,j) \in W_{xy}} L(i,j) + \min_{(i,j) \in W_{xy}} L(i,j)}
\label{eq:contrast}
\end{equation}
where $L(i,j)$ is the image's tone component, obtained by converting the colour image into the CIELAB colour space, and $W_{xy}$ is the set of co-ordinates that define the neighbourhood of $L(x,y)$. Image contrast in a grey scale image is calculated as above by substituting $I(x,y)$ for $L(x,y)$. For the comparison of contrast and agreement the maximum agreement within the local neighbourhood is used.

Ground truths at different levels of agreement are calculated such that
\begin{equation}
\gamma_\tau(x,y) = \left\{ \begin{array}{ll}
 1 & \textrm{if $\frac{1}{N} A(x,y) \ge \tau$},\\
0 & \textrm{otherwise,}
  \end{array} \right.
\label{eq:gt_threshold}
\end{equation}
where $\tau$ represents the level of annotator agreement. Additionally, a number of the gold-standard ground-truth estimation methods are evaluated. These weight annotations based upon the assumption that more reliable annotators can be identified through inter-annotator comparisons. 

To examine the inter-annotator variability, cluster analysis using the pairwise F$_1$-score between annotator markings is performed. The F$_1$-score \cite{He09}, calculated between participants $i$ and $j$, is defined as
\begin{equation}
F_{ij} = 2 \frac{p_{ij}r_{ij}}{p_{ij}+r_{ij}},
\end{equation}
and this quantity is therefore the harmonic mean of precision ($p_{ij}$) and recall ($r_{ij}$). Note that the F$_1$-score is robust in the presence of class-imbalance since it does not take into account true-negative classifications \cite{He09}. Hierarchical clustering is performed using Ward's minimum variance, implemented with the Lance-Williams dissimilarity update formula by linking pairs of annotations with the highest pair-wise F$_1$-score and repeating this until all annotations are included.

As a principled way of identifying outliers within the group of annotations, the mean F$_1$-score difference ($1 - F_{ij}$) between each annotator and all other annotators is calculated. Those that have a mean difference greater than the average plus one standard deviation are labelled as outliers.

Following the example of \citet{Saur10}, and to highlight any individual differences between the annotators, each is compared to the group's consensus (image pixels that \SI{50}{\percent} or more of the annotators marked as containing a relevant feature), calculated using Eq.\ \eqref{eq:gt_threshold} where $\tau = 0.5$. This is achieved by calculating:
\begin{description}[font=\normalfont\space,noitemsep,nolistsep]
	\item[Sensitivity,] which measures the proportion of positives that are correctly identified as such;
	\item[Specificity,] which measures the proportion of negatives that are correctly identified as such;
	\item[Positive Predictive Value (PPV),] which measures the proportions of positives that are true positives;
	\item[Negative Predictive Value (NPV),] which measures the proportions of negatives that are true negatives;
	\item[Cohen's kappa coefficient,] which measures the inter-rater agreement correcting for agreement that occurs by chance.
\end{description}


\subsection{Relation between Agreement and Detector Performance}
After analysing the properties of annotator agreement, it follows to investigate its relation to detector performance. Therefore four detectors are selected from each of the case study domains and applied to the detection problem at hand (every effort was made to select the best performing detectors within each domain). Each of the detectors is evaluated using GTs calculated at increasing levels of agreement according to Eq.\ \eqref{eq:gt_threshold}, where $\tau = 1/N, 2/N, \dots, N/N$. 

It is common to measure detector performance through ROC curve analysis, however, recent literature points out that this may overestimate performance when applied to highly skewed datasets (those in which the number of positive, $N_p$, and negative, $N_n$, examples are not balanced) and therefore precision-recall (P-R) curves are preferable \cite{Davis06, He09}. Nevertheless, precision is sensitive to the skew ratio, $\phi = {N_p} / N_n$. To overcome this \citet{Flach03} proposes to analytically vary the skew ratio in the precision measure and \citet{Lampert13b} 
 to integrate this added dimension, thus forming a $\bar{\textrm{P}}$-R curve. This allows $\bar{\textrm{P}}$-R curves derived from GTs containing different skew ratios, i.e.\ GTs derived from different levels of agreement, to be compared and for a fair representation of detector performance in problems in which the skew ratio is a priori unknown. The measure is defined such that
\begin{equation}
\bar{P}(\theta) = \frac{1}{\pi'_2 - \pi'_1} \int_{\pi'_1}^{\pi'_2} \! \frac{\pi'\textrm{TP}(\theta)}{\pi'\textrm{TP}(\theta) + (1-\pi')\phi\textrm{FP}(\theta)} \, d \pi'
\end{equation}
where $\theta$ is a threshold on the detector's output, $\pi'_1$ and $\pi'_2$ are the lower and upper bounds of the problem's estimated range of skew ratios, and TP$(\theta)$ and FP$(\theta)$ are the number of true positive and false positive detections. Interpolation between $\bar{\textrm{P}}$-R points \cite{Lampert13b} enables accurate area under the curve (AUC$\bar{\textrm{P}}$R) measurements to be taken. 

To assess the relation between annotator agreement and detector output two correlation coefficients will be measured (to \SI{99}{\percent} confidence). The first being the correlation calculated within locations identified as features by any of the annotator (CCO) and the second the whole image (CCI). The first of these highlights the relation between the detector output and annotator agreement in positive feature locations. The second includes any false positive detections that the detector may make, and therefore the absolute value of these correlations in addition to the difference between them are indicative of a detector's reliability.


\subsection{Ground Truths and Reported Detector Performance}

The final question that this research intends to investigate is: how great is the influence of different ground truths on an algorithm's reported performance?

To this end several GTs are calculated according to Eq.\ \eqref{eq:gt_threshold}: the combined annotations where $\tau = 1/N$, i.e.\ segments of interest that any annotator marked (Any-GT); the consensus of half of the annotators, or majority vote, in which $\tau = 0.5$ (0.5-GT); and the consensus of three-quarters of the annotators, where $\tau = 0.75$ (0.75-GT). Also included are gold-standard GT estimations calculated using STAPLE \cite{Warfield04} (without assigning consensus votes \cite{Commowick12}), SIMPLE \cite{Langerak10}, and LSML \cite{Li11} (using 0.5-GT as an initial estimate and \num{1000} iterations). Furthermore, an additional GT is determined by excluding outlying annotations (these will be identified in Section \ref{sec:annotator_agreement}) and combining those remaining according to Eq.\ \eqref{eq:gt_threshold} with $\tau = 0.5$ (Excl-0.5-GT).

Two forms of evaluation are investigated: the first being the relative detector ranking, ranked according to the area under the $\bar{\textrm{P}}$-R curve; and the second being the variability observed in the absolute values of the $\bar{\textrm{P}}$-R curves.


\section{Experimental Results and Analyses}
\label{sec:experiments}

This section presents the results of applying the described methodology to each of the case studies included in this investigation.


\subsection{Data}

\begin{figure*}[ht]%
	\centering%
	\subfloat[Segmentation]{
		\centering
		\includegraphics[scale=.7]{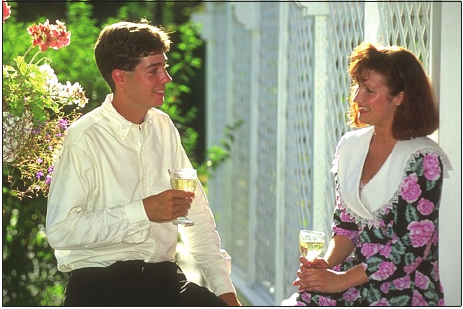}
		\label{fig:SEGMENTATION_image}
	}%
	\subfloat[Fissure]{
		\centering
		\includegraphics[scale=.7]{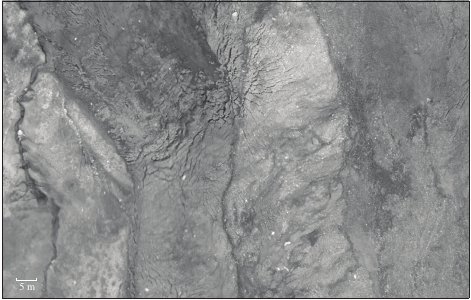}
		\label{fig:FISSURE_image}
	}\\%
	\subfloat[Landslide]{
	   \centering
		\includegraphics[scale=.7]{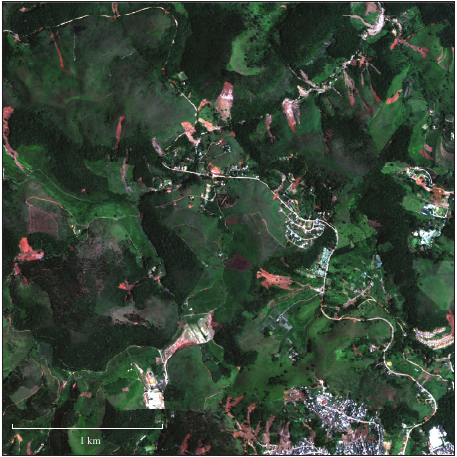}
		\label{fig:LANDSLIDE_image}
	}%
	\subfloat[Blood Vessel]{
	    \centering
		\includegraphics[scale=.7]{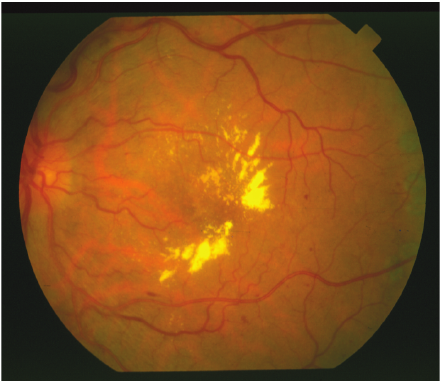}
		\label{fig:STARE_image}
	}%
	\caption{Images used in the four case studies, (a) and (d) are random examples taken from the dataset.}
\end{figure*}

The case studies presented in this section are concerned with:

\noindent\textbf{Image segmentation}
Most of the images within the Berkeley 300 (colour) dataset have been annotated by numerous different annotators. Only for a small subset of five images did the same annotators perform the segmentation (annotator IDs for the Berkeley 500 dataset are not available). These images are: 65033.jpg, 157055.jpg (Figure \ref{fig:SEGMENTATION_image}), 385039.jpg, 368016.jpg, and 105019.jpg. Each image was concatenated to form one large image, in which $X = 1595$ and $Y = 479$, and the same process was used to form one GT for each of the annotators.

\noindent\textbf{Fissures in remotely sensed images}
The data is obtained from the Super-Sauze landslide in the Barcelonnette basin, southern French Alps, using an unmanned aerial vehicle to obtain high resolution images. Further information regarding this dataset is present in the literature \cite{Niethammer11, stumpf2013image}. An area of interest, where $X = 1425$ and $Y = 906$, was extracted from the data and is presented in Figure \ref{fig:FISSURE_image}. Very little colour information is present in this type of image and it was therefore converted to grey scale using the standard formula: $I(x,y) = 0.2989 \cdot R(x,y) + 0.5870 \cdot G(x,y) + 0.1140 \cdot B(x,y)$. Thirteen annotators ($N = 13$) were enlisted to manually mark the pixels in the (RGB) image that formed part of a fissure. Within this section, each of these participants will be referred to as A1--A13. The level of expertise ranged from expert geomorphologists familiar with the study site (2), non-experts familiar with fissure formation and/or detection (5), and contributors without any \emph{a priori} knowledge (6).
Prior to the marking experiment, all the annotators were given a basic introduction to fissure characteristics. The annotators then independently marked all the pixels that they believed to form part of a fissure, taking as much time as they required (this ranged from \numrange{2}{3} hours). The annotators were encouraged to perform the marking on a level in which they could see individual pixels clearly and zoom in and out as needed to assess the context of the area being marked.

\noindent\textbf{Landslides in satellite imagery}
The dataset is derived from Geoeye-1 satellite images with four spectral bands (blue, green, red, and near infra-red) and a nominal ground resolution of \SI{50}{\cm}. The image presented in Figure \ref{fig:LANDSLIDE_image} was captured at Nova Friburgo, Brazil shortly after a major landslide event in January 2011 and covers approximately \SI{10}{\km\squared} ($X = 5960$ and $Y = 5960$ pixels). A second image was recorded by the same satellite in May 2010 and depicts the ground conditions before the event. Five annotators ($N = 5$), who were all familiar with landslide mapping in remote sensing images, were asked to independently mark the outlines of the regions affected by landslide activity. To achieve this, the RGB components of the pre-event and the post-event satellite images were visualised using a natural color scheme. Detailed information regarding this dataset exists in the literature \cite{Stumpf13}.

\noindent\textbf{Retinal blood vessels}
The STructured Analysis of the Retina (STARE) dataset was used in this case study. The dataset consists of twenty colour retinal images, which for the purposes of this study are treated as a single image ($X = 2800$ and $Y = 3025$). An example image is presented in Figure \ref{fig:STARE_image}. A mask was formed which delineates the pixels that fall outside the retina by thesholding the intensity of the red channel at a value of \num{40} (the black area) and these pixels were excluded from the experiments. The dataset contains two annotations which delineate the blood vessels in the image.

\renewcommand{\topfraction}{0.9}
\renewcommand{\dbltopfraction}{0.9}
\renewcommand{\textfraction}{0.07}

\begin{figure*}[ht]
	\centering
	\subfloat[Segmentation]{
	    \centering
		\includegraphics[scale=.70]{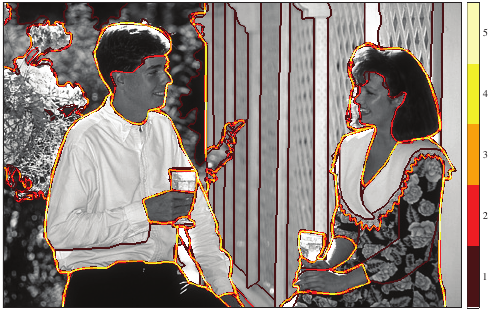}	
		\label{fig:segmentation_pixel_agreement}	
	}
	\subfloat[Fissure]{
	    \centering
		\includegraphics[scale=.70]{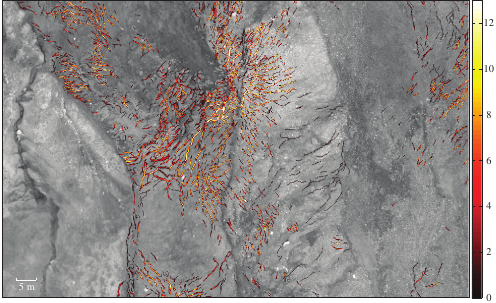}	
		\label{fig:FISSURE_pixel_agreement}
	}\\
	\subfloat[Landslide]{
	    \centering
		\includegraphics[scale=.70]{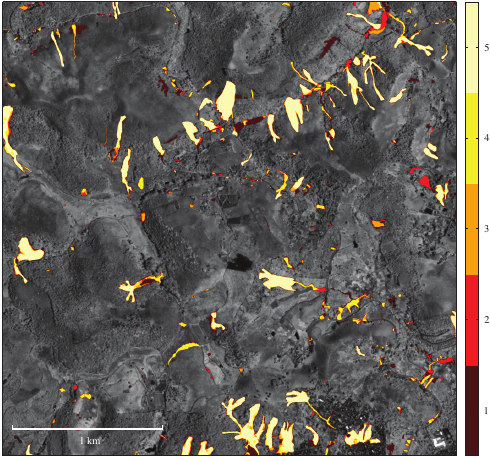}	
		\label{fig:LANDSLIDE_pixel_agreement}
	}
   	\subfloat[Blood Vessel]{
   	    \centering
   		\includegraphics[scale=.70]{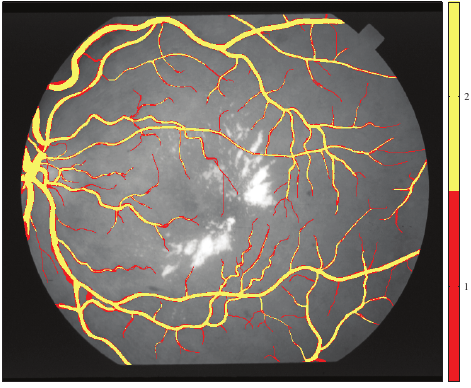}	
		\label{fig:bloodvessel_pixel_agreement}
	}
	\caption{Pixel-level annotator agreement in each case study, calculated according to Eq.\ \eqref{eq:agreement}. Colour describes the level of agreement on the location of the case study's targeted feature in the image. The images have been converted into grey-scale to better represent agreement.}
	\label{fig:AOIs}
\end{figure*}



\subsection{Annotator Agreement}
\label{sec:exp_annotator_agreement}

The pixel-level annotator agreements for each case study are presented in Figure \ref{fig:AOIs}. To verify that these are acceptable for experimental use Smyth's lower error bound estimate, i.e.\ the average error rate amongst the annotators, was calculated and found to be $\bar{e} \ge \SI{2.6611}{\percent}$ (Segmentation), $\bar{e} \ge \SI{1.26}{\percent}$ (Fissure), $\bar{e} \ge \SI{1.1012}{\percent}$ (Landslide), and $\bar{e} \ge \SI{3.1123}{\percent}$ (Blood Vessel). These values are well within the \SI{10}{\percent} limit that is recommended \cite{Smyth96} and considerably lower than the error bound of \SI{20}{\percent} found in the volcano labelling experiment presented by the author \cite{Smyth96}, in which the signal-to-noise ratio of the features is much lower than in the presented case studies.

\begin{figure}[t]
	\centering
		\includegraphics[width=0.9\linewidth]{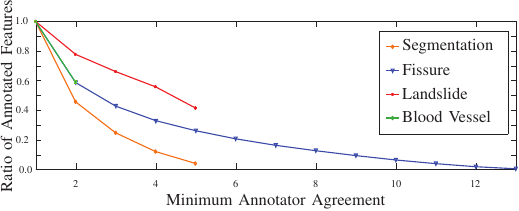}
	\caption{Ratio of the number of pixels having a minimum level of annotator agreement to the number of pixels that belong to annotated regions as the level of agreement increases, calculated according to Eq.\ \eqref{eq:agreement_marker}.}
	\label{fig:percentage_agreement}
\end{figure}

The ratio of pixels having a minimum level of annotator agreement to the number of pixels that belong to annotated regions is presented in Figure \ref{fig:percentage_agreement}. For the Segmentation, Fissure and Blood Vessel case studies the ratio decreases approximately exponentially as a function of minimum annotator agreement. For the Fissure dataset the thirteen annotators agree on only approximately \SI{0.6979}{\percent} of all of the pixels that were marked as fissures by any of the annotators. The ratio decreases most rapidly in the Segmentation case study, whereas the Landslide case study exhibits a rather linear trend. These differences are due to a combination of the geometric structure of the targeted objects, and the fact that disagreement tends to occur along object borders. As such, uncertainties in the outline of a feature lead to a stronger disagreement if the targeted features are only one pixel wide (Segmentation) or several pixels wide (Fissure, Blood Vessel) when compared to the rather blob-like regions exhibited in the Landslide case study. Indeed, if the outlines of the Landslide annotations are analysed, agreement also drops approximately exponentially.

\begin{table}[t]
	\caption{Pearson's $r$ correlation coefficients between image features and agreement. Correlations in italics are not significant at p=$0.0001$.}
	\centering
	\subfloat[Segmentation]{
		\begin{tabular}{l c c}
			\hline
			Feature & $r$\\
			\hline
			Intensity & $0.002$\\
			Contrast & $\mathbf{0.325}$\\
			Red & $-0.002$\\
			Green & $0.003$\\
			Blue & $0.033$\\
			\hline
		\label{tbl:segmentation_feature_correlations}
		\end{tabular}
		}
	\subfloat[Fissure]{
		\begin{tabular}{l c c}
			\hline
			Feature & $r$\\
			\hline
Intensity & $-0.2245$\\
Contrast & $\mathbf{0.4027}$\\
&\\
&\\
&\\
			\hline
		\label{tbl:FISSURE_feature_correlations}
		\end{tabular}
		}\\
		\subfloat[Landslide]{
		\begin{tabular}{l c c}
			\hline
			Feature & $r$ \\
			\hline
Intensity & $0.0609$\\
Contrast & $0.0310$\\
Near-IR & $\mathbf{-0.2766}$\\
Red & $0.1841$\\
Green & $-0.0115$\\
Blue & $0.0200$\\
			\hline
			\label{tbl:LANDSLIDE_feature_correlations}
		\end{tabular}
		}
		\subfloat[Blood Vessel]{
		\begin{tabular}{l c c}
			\hline
			Feature & $r$ \\
			\hline
Intensity & $-0.0861$\\
Contrast & $-0.0026$\\
Red & $0.0050$\\
Green & $\mathbf{-0.1495}$\\
Blue & \begin{tabular}[t]{@{}c@{}}$\mathit{-0.0007}$\\(p = $0.0087$)\end{tabular}\\
			\hline
			\label{tbl:bloodvessel_feature_correlations}
		\end{tabular}
		}
		\label{tbl:feature_correlations}
\end{table}

The correlation coefficients between annotator agreement and image properties are presented in Table \ref{tbl:feature_correlations}. These offer an explanation for the relation between detector performance and annotator agreement that will be explored in the remainder of this paper, i.e.\ stronger image features tend to be more confidently detected by a detection algorithm, and also attract higher levels of annotator agreement. In each case study there exists at least one significant correlation between image properties and agreement: contrast in the Segmentation case study indicating that the annotators tend to agree on stronger edges; contrast and intensity in the Fissure case study indicating that dark fissures on a lighter background attract greater agreement; near-IR and red, which exhibit a strong response if vegetation is removed during a landslide because the reddish soil is exposed; and green in the Blood Vessel case study, which is the principal channel used for discrimination in many blood vessel detection studies.

\begin{figure*}[t]
	\centering
	\subfloat[Segmentation]{
		\centering
		\includegraphics[width=0.27\linewidth]{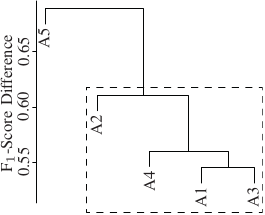}
		\label{fig:segmentation_dendrogram}
	}
	\subfloat[Fissure]{
		\centering
\includegraphics[width=0.27\linewidth]{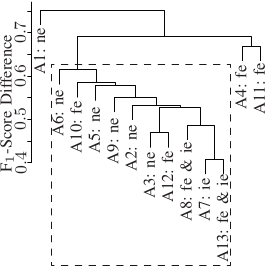}
		\label{fig:FISSURE_dendrogram}
	}
	\subfloat[Landslide]{
\includegraphics[width=0.27\linewidth]{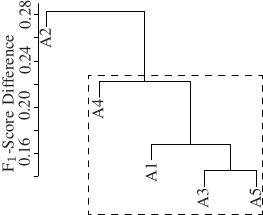}
		\label{fig:LANDSLIDE_dendrogram}
		}
		\caption{Dendrograms describing the F$_1$-score difference between each annotation for the Segmentation, Fissure, and Landslide case-studies (only two annotations are present in the Blood Vessel case study and therefore this analysis cannot be completed). Fissure case study dendrogram key: ne --- non-expert; ie --- expert with previous experience of fissure mapping in imagery; and fe --- expert with experience in the recognition of such fissures in the field. The dashed box depicts the inliers.}
		\label{fig:dendrograms}
\end{figure*}

Dendrograms describing the annotator pairwise F$_1$-scores in each case study are presented in Figure \ref{fig:dendrograms} and the full statistics of each annotator compared to the average annotation (Eq.\ \eqref{eq:gt_threshold}, $\tau = 0.5$) are presented in Table \ref{tbl:stats}.

The relatively low levels of agreement in the segmentation problem are reflected in the pairwise differences in F$_1$-scores used to form the dendrogram in Figure \ref{fig:segmentation_dendrogram}. The differences are relatively high, ranging from \num{0.545} to \num{0.680}, and one outlier is identified: A5 (the mean F$_1$-score difference was found to be \num{0.6016}, with a standard deviation of \num{0.0280}, and A5 resulted in a difference of \num{0.6454}). This annotator also results in the lowest specificity, positive predictive value, and kappa coefficient as shown in Table \ref{tbl:segmentation_stats}. The variance in the annotations are emphasised by the lowest specificities observed in all of the case studies. A dendrogram describing the Blood Vessel dataset is not included as no outliers can be identified with only two annotations. Nevertheless, the F$_1$-score difference ($1-F_{ij}$) calculated between the two annotations was found to be \num{0.2583} meaning that they give fairly consistent markings. The statistics in Table \ref{tbl:bloodvessel_stats} are not as informative as in the other case studies due to the low number of annotators and this highlights one of the issues of estimating GTs using few annotations and such statistical comparisons. Nevertheless, we can infer that A2 marked a much larger number of blood vessels compared to A1 due to A2 having a high sensitivity and A1 not (in this case the \SI{50}{\percent} agreement GT with which these statistics are calculated contains locations that any of the annotators marked, hence the specificity and PPV being one).

It would be expected that more than one cluster emerges within the Fissure case study, Figure \ref{fig:FISSURE_dendrogram}, partitioning the different experience levels; however, this isn't the case and annotators of varying levels of expertise are quite homogeneously mixed. This indicates that none of the groups is overly biased in favour of one particular decision. Annotators A1, A4, and A11 are identified as falling outside of one standard deviation of the mean F$_1$-score difference. These same annotators achieve considerably lower sensitivity when compared to the consensus (see Table \ref{tbl:FISSURE_stats}). They also result in lower kappa coefficients and PPVs---indicating that, when compared to the consensus, these annotators fail to identify a majority of the fissures and/or produce more `false negative' and `false positive' detections. The mean F$_1$-score difference ($1-F_{ij}$) is found to be $0.5765$ and the standard deviation \num{0.0459}, these annotators fall outside this threshold having a mean difference of \num{0.6716}, \num{0.6321}, and \num{0.6287} (corresponding to A1, A4, and A11 respectively). It is illustrated by these results that all of the annotators are reliable in detecting negative instances of fissures, indicated by high specificity and negative predictive values, due to the highly skewed nature of the problem in which negative instances constitute a high proportion of the data. Highlighting the difficulty and uncertainty in detecting positive instances in this dataset, however, are low sensitivity and PPVs.

In the Landslide case study, each of the annotators were geographers familiar with the detection of landslides in remotely sensed imagery. This is reflected in the low inter F$_1$-score difference ($1-F_{ij}$), which ranges from \num{0.14} to \num{0.28} (by comparison this range was approximately \num{0.40} to \num{0.75} in the Fissure case study). Nevertheless, one outlier is identified and this is A2 (the mean difference was found to be \num{0.2044} and its standard deviation \num{0.0275}, A2 resulted in a mean difference of \num{0.2438}). This annotator also results in the lowest of the sensitivity and negative predictive values (when compared to the consensus opinion) presented in Table \ref{tbl:LANDSLIDE_stats}. On average, sensitivity, PPV and kappa are higher than in the Fissure case study, indicating that the features used for the identification of landslides are more clearly defined and understood by the annotators.

\begin{table}[htp]
	\caption{Sensitivity (Sens.), specificity (Spec.), positive predictive value (PPV), negative predictive value (NPV) and Cohen's kappa coefficient of the participants when compared to the consensus (rounded to four decimal places).}
	\centering
	\subfloat[Segmentation]{
    \begin{tabular}{l c c c c c}
		\hline
		& Sens. & Spec. & PPV & NPV & kappa \\
		\hline
A1 & $0.7694$ & $0.9845$ & $0.5634$ & $0.9939$ & $0.6399$\\
A2 & $0.6373$ & $0.9886$ & $0.5921$ & $0.9905$ & $0.6034$\\
A3 & $0.7853$ & $0.9785$ & $0.4882$ & $0.9943$ & $0.5892$\\
A4 & $0.7309$ & $0.9822$ & $0.5166$ & $0.9929$ & $0.5933$\\
\textbf{A5} & $0.7275$ & $\mathbf{0.9649}$ & $\mathbf{0.3509}$ & $0.9927$ & $\mathbf{0.4548}$\\
		\hline
		\label{tbl:segmentation_stats}
	\end{tabular}	
	}\\
	\subfloat[Fissure]{
		\begin{tabular}{l c c c c c}
			\hline
			& Sens. & Spec. & PPV & NPV & kappa \\
			\hline
\textbf{A1} & $\mathbf{0.5595}$ & $0.9847$ & $\mathbf{0.2893}$ & $0.9950$ & $\mathbf{0.3722}$\\
A2 & $0.7518$ & $0.9911$ & $0.4860$ & $0.9972$ & $0.5848$\\
A3 & $0.7526$ & $0.9945$ & $0.6018$ & $0.9972$ & $0.6647$\\
\textbf{A4} & $\mathbf{0.5705}$ & $0.9906$ & $\mathbf{0.4032}$ & $0.9952$ & $\mathbf{0.4656}$\\
A5 & $0.6429$ & $0.9938$ & $0.5362$ & $0.9960$ & $0.5797$\\
A6 & $0.6244$ & $0.9926$ & $0.4834$ & $0.9958$ & $0.5392$\\
A7 & $0.9380$ & $0.9866$ & $0.4377$ & $0.9993$ & $0.5907$\\
A8 & $0.7897$ & $0.9906$ & $0.4828$ & $0.9976$ & $0.5937$\\
A9 & $0.6894$ & $0.9926$ & $0.5106$ & $0.9965$ & $0.5814$\\
A10 & $0.6659$ & $0.9925$ & $0.4969$ & $0.9963$ & $0.5636$\\
\textbf{A11} & $\mathbf{0.5799}$ & $0.9899$ & $\mathbf{0.3905}$ & $0.9953$ & $\mathbf{0.4596}$\\
A12 & $0.7461$ & $0.9937$ & $0.5672$ & $0.9972$ & $0.6399$\\
A13 & $0.8738$ & $0.9836$ & $0.3719$ & $0.9986$ & $0.5143$\\
			\hline
			\label{tbl:FISSURE_stats}
		\end{tabular}
		}\\
		\subfloat[Landslide]{
		\begin{tabular}{l c c c c c}
			\hline
			& Sens. & Spec. & PPV & NPV & kappa \\
			\hline
A1 & $0.9280$ & $0.9942$ & $0.8837$ & $0.9966$ & $0.9007$\\
\textbf{A2} & $\mathbf{0.7499}$ & $0.9972$ & $0.9276$ & $0.9883$ & $\mathbf{0.8222}$\\
A3 & $0.8797$ & $0.9978$ & $0.9502$ & $0.9943$ & $0.9097$\\
A4 & $0.9713$ & $0.9837$ & $0.7380$ & $0.9986$ & $0.8300$\\
A5 & $0.9419$ & $0.9945$ & $0.8893$ & $0.9972$ & $0.9107$\\
			\hline
			\label{tbl:LANDSLIDE_stats}
		\end{tabular}
		}\\
		\subfloat[Blood Vessel]{
		\begin{tabular}{l c c c c c}
			\hline
			& Sens. & Spec. & PPV & NPV & kappa \\
			\hline
A1 & $0.6536$ & $1.0000$ & $1.0000$ & $0.9417$ & $0.4956$\\
A2 & $0.9358$ & $1.0000$ & $1.0000$ & $0.9887$ & $0.5702$\\
			\hline
			\label{tbl:bloodvessel_stats}
		\end{tabular}
		}
		\label{tbl:stats}
\end{table}


\subsection{Agreement and Detector Performance}

During these case studies a number of detectors were selected and their ability to detect features in the area of interest was evaluated by calculating $\bar{\textrm{P}}$-R curves:

\noindent\textbf{Segmentation}
The top four performing segmentation algorithms listed on the Berkeley dataset web page\footnote{\url{http://www.eecs.berkeley.edu/Research/Projects/CS/vision/grouping/segbench/bench/html/algorithms.html}, accessed \nth{23} October 2015} were selected to form part of this case study. These were: REN \cite{Ren12}, gPb-ucm (UCM) \cite{Arbelaez11}, Global Probability of Boundary (GP) \cite{Maire08}, and XREN \cite{Ren08}. The integration limits of the $\bar{\textrm{P}}$-R curves were $\pi'_1 = 0.0000$ and $\pi'_2 = 0.0428$, which were found to be $\pi'_1 = \mu - 3\sigma$ and $\pi'_2 = \mu + 3\sigma$ where $\mu$ is the mean skew found within the Berkeley dataset and $\sigma$ its standard deviation \cite{Lampert13b}. As discussed by \citet{Martin04}, when evaluating segmentation algorithms it is common to loosen the definition of true-positive detections to account for deviations in detected boundary location. True-positive detections are accumulated if a detection is within a defined distance of one or more GT boundaries. In these experiments the allowed distance is taken to be the default found with the Berkeley benchmark code---\num{0.0075} times the length of the image's diagonal. The images 105019.jpg and 368016.jpg are randomly selected for use as the training set and removed from this point forward. One further modification to the methodology was made to better suite the definition of segmentation. The low agreement GTs ($\tau = 1 / N$, for example) result in multiple pixel wide segmentations (as annotators may agree upon the boundary's existence, but not on its exact location), which causes an unfair penalty on the algorithm because a segmentation algorithm is designed to detect single pixel segmentation boundaries. Therefore, each GT is thinned prior to its use to reduce the boundary widths to one pixel whilst preserving any individual, low agreement markings.

\noindent\textbf{Fissure}
Current state-of-the-art linear feature detectors were selected from the literature: a linear classifier trained using 2D Gabor wavelet (elongation $\epsilon = 4$, scales $a = 2, 3, 4, 5$, and frequency $k_0 = 3$) and inverted grey-scale features (2D GWLC) \cite{Soares06}; Gaussian filter matching, where $\sigma = 1$ \cite{Stumpf12} (Gauss); Top-Hat transform ($4$ pixel radius circular structuring element); and the Centre-Surround (C-S) transform (using a $3 \times 3$ pixel neighbourhood) \cite{Vonikakis08}. Where public source code was not available the respective authors kindly agreed to run the algorithm on the data and provide a number of outputs, calculated using a range of parameter values (to ensure that the implementations were true to the author's intentions and to allow reproducibility of the results). As the 2D GWLC method is a supervised learning algorithm a random subset of the image, $569 \times 362$ pixels in size, was used as a training set (\SI{16}{\percent} of the image), the GT was defined according to Eq.\ \eqref{eq:gt_threshold} using $\tau = 1/N$, and the training area was excluded from the test set. Within this case study the $\bar{\textrm{P}}$-R integration limits were set to $\pi'_1 = 0.1$ and $\pi'_2 = 0.5$ (from ten times as many negative as positive instances to a balanced dataset) to reflect the large range of skews that can be observed in a remote sensing application.

\noindent\textbf{Landslide}
Four popular classification algorithms were applied (due to their proven strength in real-world applications): random forest (RF) \cite{Liaw02}, support-vector machine (SVM), $k$-nearest neighbours (KNN), and a neural network (ANN).
After fine scale image segmentation, \num{101} features describing the spectral characteristics, texture, shape, topographic variables, and neighbourhood contrast were extracted. The resulting dataset is available on-line\footnote{\url{http://eost.unistra.fr/recherche/ipgs/dgda/dgda-perso/andre-stumpf/data-and-code/}} and a detailed description of the feature extraction methods are given in the literature \cite{Stumpf13}. Each classifier was trained upon samples from the same randomly selected square subset covering \SI{10}{\percent} of the area of interest. The number of trees in the RF was fixed to \num{500} and \num{10} variables were tested for the splits at each node. The SVM used a radial basis kernel having parameters $C=10$ and $\sigma=0.004$, determined through an exhaustive grid search. The ANN was a single layer network with a logistic activation function. An exhaustive grid search to optimize the weight decay function and the number of nodes resulted in values of \num{0.1} and \num{7}, respectively. Likewise, a grid search for the number of nearest neighbours resulted in $k=23$ for the KNN algorithm. Parameter tuning was performed through bootstrap resampling of the training data using the area under the ROC curve as a performance measure. The $\bar{\textrm{P}}$-R integration limits were set to $\pi'_1=0.01$ and $\pi'_2=0.10$ to reflect typical ratios of affected to unaffected areas after large-scale landslide triggering events \cite{Malamud04}. 

\noindent\textbf{Blood Vessel}
The four detectors selected for this case study were the Matched-Filter Response (MSF) \cite{Hoover00}, Linear Classifier (LMSE), $k$-nearest neighbours (KNN), and Gaussian Mixture Model (GMM). The LMSE, KNN and GMM classifiers were implemented using the MLVessel software package \cite{Soares06}, the features were taken to be the inverted green channel, and the responses of Gabor wavelets (elongation $\epsilon = 4$, scales $a = 2, 3, 4, 5$, and frequency $k_0 = 3$) applied to the inverted green channel. The first five images of the dataset (im0001--5) were used exclusively for training. The $\bar{\textrm{P}}$-R curve integration limits were $\pi'_1 = 0.023$ and $\pi'_2 = 0.235$, which were found to be $\pi'_1 = \mu - 3\sigma$ and $\pi'_2 = \mu + 3\sigma$ where $\mu$ was the mean skew found within a number of retinal image datasets and $\sigma$ its standard deviation \cite{Lampert13b}.

\begin{figure*}[htp]
	\centering
	\subfloat[GP]{
		\centering
\includegraphics[width=0.202\textwidth]{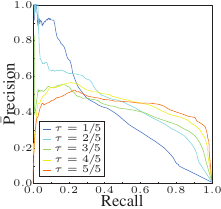}
		\label{fig:roc_gp}
	}
	\subfloat[REN]{
		\centering
\includegraphics[width=0.19\textwidth]{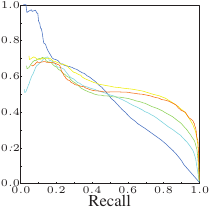}
		\label{fig:roc_ren}
	}
 	\subfloat[UCM]{
		\centering      
\includegraphics[width=0.19\textwidth]{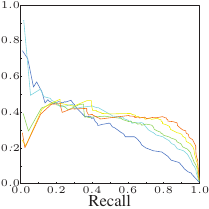}
		\label{fig:roc_ucm}
	}
	\subfloat[XREN]{
		\centering
\includegraphics[width=0.19\textwidth]{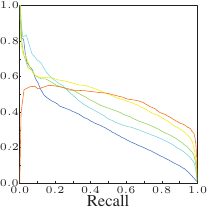}
		\label{fig:roc_xren}
	}\\	
	\subfloat[2D GWLC]{
		\centering
\includegraphics[width=0.202\textwidth]{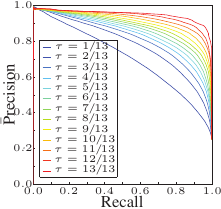}
		\label{fig:roc_blood}
	}
	\subfloat[Single Scale Gaussian]{
		\centering
\includegraphics[width=0.19\textwidth]{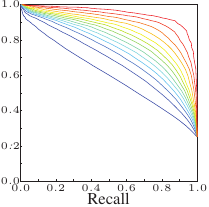}
		\label{fig:roc_single_scale_guass}
	}
	\subfloat[Top-Hat]{
		\centering
\includegraphics[width=0.19\textwidth]{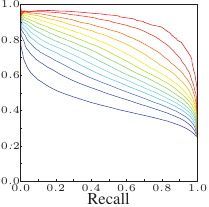}
    		\label{fig:roc_top_hat}
    }
	\subfloat[Centre-Surround]{
		\centering
\includegraphics[width=0.19\textwidth]{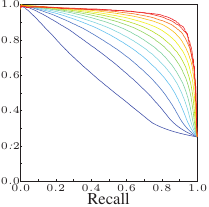}
        \label{fig:roc_centre_surround}
	}\\
		\subfloat[RF]{
		\centering
\includegraphics[width=0.202\textwidth]{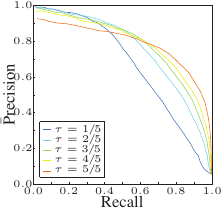}
		\label{fig:roc_rf}
	}
	\subfloat[SVM]{
		\centering
\includegraphics[width=0.19\textwidth]{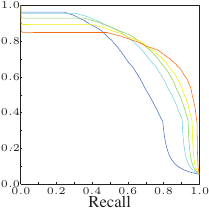}
		\label{fig:roc_svm}
	}
	\subfloat[KNN]{
		\centering
\includegraphics[width=0.19\textwidth]{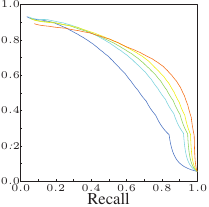}
		\label{fig:roc_knn_landslide}
	}
	\subfloat[ANN]{
		\centering
\includegraphics[width=0.19\textwidth]{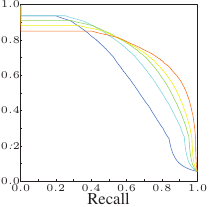}
		\label{fig:roc_ann}
	}\\
	\subfloat[MSF]{
		\centering
\includegraphics[width=0.202\textwidth]{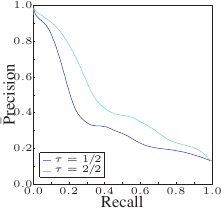}
		\label{fig:roc_msf}
	}
	\subfloat[LMSE]{
		\centering
\includegraphics[width=0.19\textwidth]{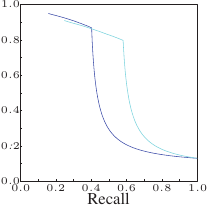}
		\label{fig:roc_lmse}
	}
	\subfloat[KNN]{
		\centering
\includegraphics[width=0.19\textwidth]{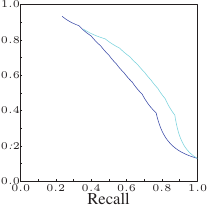}
		\label{fig:roc_knn_stare}
	}\hspace{-1.5em}
	\subfloat[GMM]{
		\centering
\includegraphics[width=0.19\textwidth]{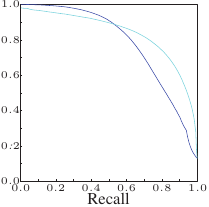}
		\label{fig:roc_gmm}
	}
	\caption{$\bar{\textrm{P}}$recision-recall curves describing detector performance. The curves in each subfigure are determined using ground truths calculated with increasing levels of agreement, according to Eq.\ \eqref{eq:gt_threshold}. Figures \ref{fig:roc_gp}--\ref{fig:roc_xren} are from the Segmentation case study, Figures \ref{fig:roc_blood}--\ref{fig:roc_centre_surround} from the Fissure case study, Figures \ref{fig:roc_rf}--\ref{fig:roc_ann} from the Landslide case study, and Figures \ref{fig:roc_msf}--\ref{fig:roc_gmm} from the Blood Vessel case study.}
	\label{fig:ipr_curves}
\end{figure*}

The $\bar{\textrm{P}}$-R curves derived from these detectors are presented in Figure \ref{fig:ipr_curves}. A striking observation is that the performance of all detectors increases with annotator agreement in a predictable manner in the higher recall ranges. It was shown in Table \ref{tbl:feature_correlations} that there is a tenancy for annotators to agree upon more obvious image features, and these results indicate that the detectors extract similar features. Regarding the Fissure dataset, there is a large difference between the detection rate of high and low agreement fissures---detection of the lower is not a trivial matter and most likely needs to be augmented with high-level information that is not exploited by the evaluated detectors. In the lower recall ranges of the Segmentation, Landslide, and Blood Vessel case studies the tendency for precision to increase with agreement is reversed. This phenomenon can be explained by analysing the correlations between annotator agreement and detector output presented in Table \ref{tbl:FISSURE_detector_correlations}.

Several general tendencies can be drawn from these correlations. The detectors that exhibit a large drop between CCO and CCI also exhibit low sensitivity (i.e.\ produce a high false-positive rate). For example, this is reflected in the $\bar{\textrm{P}}$-R curves of the C-S detector (Figure \ref{fig:roc_centre_surround}): low sensitivity dominates the low agreement ground truths (for example, $\tau=1/13$), but the detector results in the highest performance when using the high agreement ground truths (for example, $\tau=1$). The detectors that exhibit a high correlation with agreement over the whole image, and also exhibit the lowest drop in correlation between the two tests (2D GWLC, Gauss, SVM, \& GMM detectors for example), have (relatively) low false positive rates and result in high $\bar{\textrm{P}}$-R curves (Figures \ref{fig:roc_blood}, \ref{fig:roc_single_scale_guass}, \ref{fig:roc_svm}, \& \ref{fig:roc_gmm}). A large drop in correlation, along with a low absolute correlation, is observed with the Top-Hat detector, and indeed in Fig.\ \ref{fig:roc_top_hat} the curves are skewed towards lower precision values. The detectors that result in the lowest drop or an increase in correlation (2D GWLC, Gauss, RF, KNN, SVM, \& ANN) result in a tighter spread of $\bar{\textrm{P}}$-R curves (Figures \ref{fig:roc_blood}, \ref{fig:roc_single_scale_guass}, \ref{fig:roc_rf}, \ref{fig:roc_knn_landslide}, \ref{fig:roc_svm}, \& \ref{fig:roc_ann}).

The $\bar{\textrm{P}}$-R curves from the Segmentation, Landslide and Blood Vessel case studies largely follow the trend: as agreement increases, algorithm performance also increases. There is, however, a tendency for precision to be inversely proportional to agreement in lower recall ranges. This phenomenon can be explained by analysing the correlations between annotator agreement and detector outputs presented in Table \ref{tbl:FISSURE_detector_correlations} and noting that in all the cases in which this trend is observed CCI is higher than CCO. This indicates that detector outputs agree with annotator agreement within feature locations and more so over the whole image, implying that there is a relatively low FP detection rate, which at the lower recall ranges results in high precision. As the threshold on agreement increases, image locations having increasingly stronger features form the GT and these locations also result in the highest detector responses. Furthermore, high CCI values imply that as lower agreement segments are removed from the GT they are instead classified as false positive detections, thus reducing precision in the lower recall ranges as the threshold on annotator agreement is increased.

\begin{table}[t]
	\caption{Pearson's $r$ correlation coefficients between detector outputs and annotator agreement $A(x,y)$ as defined by Eq.\ \eqref{eq:agreement}; CCO is calculated within the pixels marked as a positive instance by any of the annotators, and CCI the whole image. The p-values are all \num{0.0000} (to four decimal places).}
	\centering
		\begin{tabular}{c l c c c}
			\hline
			Case Study & Detector & CCO & CCI & CCI$-$CCO\\
			\hline
\multirow{4}{*}{Segmentation} & UCM & $\mathbf{0.2686}$ & $\mathbf{0.3663}$ & $+0.0977$\\
& GP & $0.1603$ & $0.2746$ & $\mathbf{+0.1143}$\\
& XREN & $0.2633$ & $0.3206$ & $+0.0573$\\
& REN & $0.2089$ & $0.3119$ & $+0.1030$\\
\hline
\multirow{4}{*}{Fissure} & 2D GWLC & $0.5563$ & $0.5166$ & $\mathbf{-0.0397}$\\
& Gauss & $0.5293$ & $0.4711$ & $-0.0582$\\
& C-S & $\mathbf{0.6387}$ & $\mathbf{0.5259}$ & $-0.1128$\\
& Top-Hat & $0.5187$ & $0.2780$ & $-0.2407$\\
\hline
\multirow{4}{*}{Landslide} & RF & $0.6497$ & $0.7829$ & $+0.1332$\\
& KNN & $0.6072$ & $0.7551$ & $+0.1479$\\
& SVM & $\mathbf{0.6503}$ & $\mathbf{0.7992}$ & $\mathbf{+0.1489}$\\
& ANN & $0.6417$ & $0.7565$ & $+0.1148$\\
\hline
\multirow{4}{*}{Blood Vessel} & MSF & $0.3923$ & $0.3573$ & $-0.0350$\\
& GMM & $\mathbf{0.5833}$ & $\mathbf{0.8133}$ & $\mathbf{+0.2300}$\\
& LMSE & $0.4168$ & $0.5950$ & $+0.1782$\\
& KNN & $0.4361$ & $0.6952$ & $+0.2591$\\
			\hline
			\label{tbl:FISSURE_detector_correlations}
		\end{tabular}
		\label{tbl:detector_correlations}
\end{table}


\subsection{Ground Truths and Reported Detector Performance}

The detector ranks (measured as AUC$\bar{\textrm{P}}$R \cite{Lampert13b}) when evaluated using different GTs were determined, and in three of the case studies (Segmentation, Fissure, and Blood Vessel) three rankings emerged, which are described in Table \ref{tbl:detector_ranks}. In the Landslide case study only one emerged due to the low inter-annotator variance. In the Fissure, Landslide, and Blood Vessel case studies these ranks reflect the results of the correlation analyses: the top ranked detectors (2D GWLC, SVM, and GMM) and the bottom ranked detectors (Top-Hat and KNN) correspond to either the highest correlations or the lowest drops in correlation observed in the previous section (see Table \ref{tbl:detector_correlations}). Furthermore, in the Fissure case study the ranking observed is not determined by the level of annotator expertise. This corroborates the lack of distinction between different expertise levels in the dendrogram presented in Figure~\ref{fig:FISSURE_dendrogram}.

In the Segmentation case study, however, the algorithms with the highest correlation (UCM) and the highest CCO to CCI increase (GP) are ranked at the middle or bottom. On one hand this can be attributed to the relatively high annotation variance and the overall low correlation between detector output and the annotator agreement (Table \ref{tbl:detector_correlations}). On the other hand it should also be considered that the correlations are derived using all of the annotated pixels, while the $\bar{\textrm{P}}$-R curves are calculated using GTs that were thinned to a width of one pixel and the TP rates calculated with a tolerance to small mismatches of the segmentation boundary. This not only highlights the sensitivity of evaluating algorithms using different GTs that exhibit high variance, but also illustrates how different evaluation strategies can provoke different outcomes.

\begin{figure}[t]
	\centering
\includegraphics[width=0.68\linewidth]{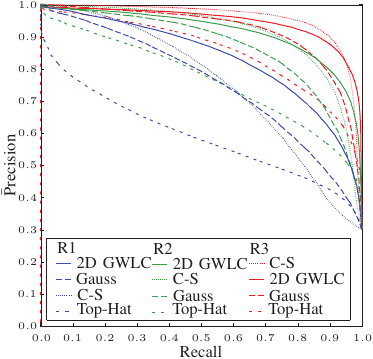}
	\caption{$\bar{\textrm{P}}$recision-recall curves of all four detectors in the Fissure detection case study evaluated using: STAPLE-GT which results in Ranking \#1 (R1); 0.5-GT which results in Ranking \#2 (R2); and 0.75-GT which results in Ranking \#3 (R3).}
	\label{fig:FISSURE_ranking_ipr_curves}
\end{figure}

In the Fissure case study, a majority of the individual annotations give the same ranking as obtained using the SIMPLE-GT, and 0.5-GT, however, when the 0.75-GT, Any-GT, STAPLE-GT, and LSML-GT are under consideration, the ranking changes---the method of calculating the GT influences detector ranking. More importantly, the ranking derived using a \SI{75}{\percent} voting strategy (Fissure) and LSML (Blood Vessel) are in disagreement with that obtained using the individual annotations, which contradicts what should be expected. To illustrate ranks in the Fissure case study, the $\bar{\textrm{P}}$-R curves for all four detectors evaluated using the STAPLE-GT, 0.5-GT, and 0.75-GT are plotted in Figure \ref{fig:FISSURE_ranking_ipr_curves}, each colour represents one of the rankings presented in Table IVb.

In the Blood Vessel case study the ranks of the lower three detectors are not consistent. The MSF detector, for example, achieves the lowest performance in Figure \ref{fig:ipr_curves} along with the lowest correlation with annotator agreement (Table \ref{tbl:detector_correlations}), however, depending upon which GT is used, this detector is placed second, third, or last.

\captionsetup[subfigure]{position=b}
\begin{table}[t]
	\caption{Rankings of detectors evaluated using each ground truth (measured by the area under the $\bar{\textrm{P}}$-R curve). (a) Segmentation case study, the GTs that result in each ranking are: Ranking \#1 --- Berkeley evaluation framework (A1--A5); Ranking \#2 --- A4, A5, Any-GT, LSML-GT, STAPLE-GT; Ranking \#3 --- A1, A2, A3, 0.5-GT, 0.75-GT, Excl-0.5-GT, SIMPLE-GT. (b) Fissure case study, the GTs that result in each ranking are: Ranking \#1 --- A2(ne), A4(fe), A6(ne), A11(fe), Any-GT, LSML-GT, STAPLE-GT, Excl-0.5-GT; Ranking \#2 --- A1(ne), A3(ne), A5(ne), A7(ie), A8(fe \& ie), A9 (ne), A10(fe), A12(fe), A13(fe \& ie), 0.5-GT, SIMPLE-GT; Ranking \#3 --- 0.75-GT. (c) Landslide case study, all GTs result in the same ranking. (d) Blood Vessel case study, the GTs that result in each ranking are: Ranking \#1 --- A1, 0.75-GT, SIMPLE-GT; Ranking \#2 --- A2, 0.5-GT/Any-GT, STAPLE-GT; Ranking \#3 --- LSML-GT.}
	\centering
	\subfloat[Segmentation]{
	\begin{tabular}{c c c c}
		\hline
Position & Ranking \#1 & Ranking \#2 & Ranking \#3\\
\hline
1 & REN & REN & REN\\
2 & GP & GP & XREN\\
3 & UCM & XREN & GP\\
4 & XREN & UCM & UCM\\
		\hline
		\label{tbl:segmentation_detector_ranks}
	\end{tabular}
	    }\\
	\subfloat[Fissure]{
		\begin{tabular}{c c c c}
			\hline
Position & Ranking \#1 & Ranking \#2 & Ranking \#3\\
\hline
1 & 2D GWLC & 2D GWLC & C-S\\
2 & Gauss & C-S & 2D GWLC\\
3 & C-S & Gauss & Gauss\\
4 & Top-Hat & Top-Hat & Top-Hat\\
			\hline
			\label{tbl:FISSURE_detector_ranks}
		\end{tabular}
		}\\
		\subfloat[Landslide]{
		\begin{tabular}{c c}
			\hline
Position & Ranking \#1\\
\hline
1 & SVM\\
2 & RF\\
3 & ANN\\
4 & KNN\\
			\hline
			\label{tbl:LANDSLIDE_detector_ranks}
		\end{tabular}
		}\\
		\subfloat[Blood Vessel]{
		\begin{tabular}{c c c c c}
			\hline
Position & Ranking \#1 & Ranking \#2 & Ranking \#3\\
\hline
1 & GMM & GMM & GMM\\
2 & MSF & KNN & KNN\\
3 & LMSE & MSF & LMSE\\
4 & KNN & LMSE & MSF\\
			\hline
			\label{tbl:bloodvessel_detector_ranks}
		\end{tabular}
		}
		\label{tbl:detector_ranks}
\end{table}

\begin{figure*}[htp]
	\centering
	\subfloat[Segmentation]{
\includegraphics[width=0.28\textwidth]{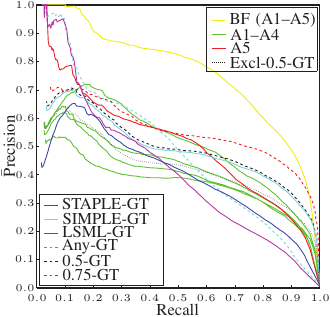}
	\label{fig:segmentation_gt_ipr_curves}
	}
	\subfloat[Fissure]{
\includegraphics[width=0.28\textwidth]{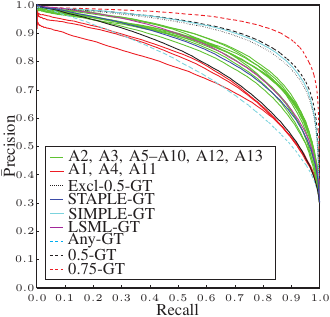}
	\label{fig:FISSURE_gt_ipr_curves}
	}\\
	\subfloat[Landslide]{
\includegraphics[width=0.28\textwidth]{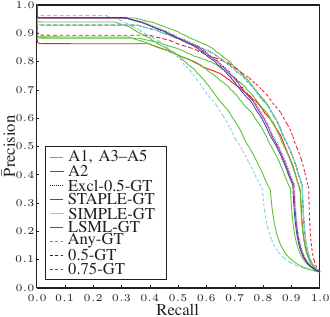}
	\label{fig:LANDSLIDE_gt_ipr_curves}
	}
	\subfloat[Blood Vessel]{
\includegraphics[width=0.28\textwidth]{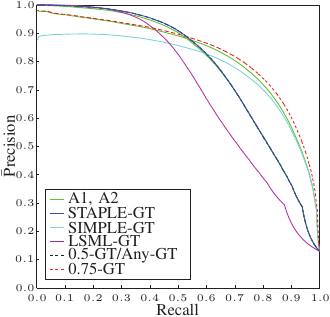}
	\label{fig:bloodvessel_gt_ipr_curves}
	}
	\caption{$\bar{\textrm{P}}$recision-recall curves for one detector in each case study using different ground truth estimation methods: (a) the REN detector, (b) the 2D GWLC detector (c), RF (0.5-GT and SIMPLE-GT are identical) (d), GMM (the curve obtained using STAPLE-GT overlaps that obtained using 0.5-GT and Any-GT, 0.5-GT, and Excl-0.5-GT are identical).}
	\label{fig:gt_ipr_curves}
\end{figure*}

An overview of the performance variations that result from using different GT estimation methods and evaluation frameworks can be obtained from Figure \ref{fig:gt_ipr_curves}, in which the $\bar{\textrm{P}}$-R curves obtained using the best performing detector in each of the case studies are presented. The $\bar{\textrm{P}}$-R curves for the REN segmentation algorithm (Figure \ref{fig:segmentation_gt_ipr_curves}) exhibit the largest level of variance, a result of the high annotator variance observed in Section \ref{sec:annotator_agreement}. At the upper extreme of this variance is the methodology prescribed for evaluating segmentation algorithms upon the Berkeley datasets, which includes a tolerance for misalignments of TP detections. The 0.75-GT, 0.5-GT, and SIMPLE-GT yield higher performance curves (particularly in higher recall ranges) and Any-GT relatively low performance curves when compared to the remaining GTs. The STAPLE-GT and LSML-GT curves show low $\bar{\text{p}}$recision (when compared to the remaining curves) in the upper recall ranges, but model the mean of the individual annotation curves in the lower recall ranges. This is a consequence of the large variance observed in the annotations. The curves resulting from Excl-0.5-GT and SIMPLE-GT are very similar as they are both derived using the same principle (removing outliers and then voting).

The $\bar{\textrm{P}}$-R curves resulting from 2D GWLC are presented in Figure \ref{fig:FISSURE_gt_ipr_curves}. The effects of the voted GTs (0.5-GT, 0.75-GT, and SIMPLE-GT) become evident: these $\bar{\textrm{P}}$-R curves estimate a relatively high detector performance and seem to act as generous estimates of the upper bound of the performance derived from the individual annotations. Moreover, the Any-GT appears to act as an estimate of the lower bound of the performance derived from the individual annotations, and when sufficient annotations are available (Fissure and Landslide) the curves obtained using STAPLE-GT and LSML-GT appear to approximately model the mean of the performance obtained using the inlying individual annotations. It should be noted, however, that the LSML technique is highly dependent upon the estimate used for initialisation.

Similarly, in the Landslide case study (Figure \ref{fig:LANDSLIDE_gt_ipr_curves}) 0.5-GT, 0.75-GT, and SIMPLE-GT yield $\bar{\textrm{P}}$-R curves that seem to model the upper bound of the performance obtained using the individual annotations, STAPLE and LSML tend to produce GTs that result in $\bar{\textrm{P}}$-R curves that are within the range of those obtained using the individual annotations, and Any-GT marks the lower bound of the detector's performance. Overall it can be observed that the lower annotator variance observed in this case study leads to a significantly lower $\bar{\textrm{P}}$-R curve spread.

On the contrary, in the Blood Vessel case study (Figure \ref{fig:bloodvessel_gt_ipr_curves}, due to the limited number of annotations the Any-GT, 0.5-GT, and Excl-0.5-GT are identical) the LSML-GT forms a lower bound on the reported performance. The STAPLE-GT (equal to the 0.5-GT and the Any-GT) delineates the mean of all the curves, whereas previously (but to a lesser extent in the Segmentation case study) the STAPLE-GT and LSML-GT represented an estimate of the mean of the curves obtained using the individual annotations. Once more 0.75-GT results in a higher estimate of performance than that obtained using each of the individual annotations.

\section{Discussion}
\label{sec:discussion}

The following discussion is divided into two parts: the first summarises the results presented in the previous section and their implications, and the second presents general recommendations that can be derived from these implications.


\subsection{Summary of Results}

It has been shown that the performance of classifiers and detectors increases as GTs are formed using increasingly higher agreement levels. Forming a GT using an agreement of \SI{50}{\percent} generally increases a detector's reported performance to a range far greater than that obtained using all of the individual annotations. \citet{Kauppi09} conclude that the intersection method (consensus) is preferential as it results in the highest performance. Nevertheless, this study indicates that the method focusses on evaluating a detector against the most obvious segments in the image and provides overly optimistic performance estimates. Raising the level of agreement at which the GT is calculated increases this tendency.

One factor that has a stabilising effect on reported performance is low annotation variance. The Landslide dataset contains the lowest variance between annotations and this is reflected in the tight spread of the performance curves and in the stability of the detector ranking. Hence choosing any of the GTs for evaluating an algorithm would have resulted in similar reported performance. On the other end of the scale the Segmentation dataset contained the largest annotation variance, and the reported performances also exhibit the largest variance. This is in contrast to the findings of \citet{Martin01} who found a large amount of agreement between the segmented regions, but not the boundaries themselves. This also affected the gold-standard GT estimation methods, where in the other case studies the STAPLE and LSML methods typically modelled the `mean' performance derived using the individual annotations, in this case study they actually resulted in the lowest performance curves. Both of these methods combine annotations using the annotator's statistical profile and given that there is a large variance in this dataset this may not be appropriate. In this situation removing the outlier annotations and performing consensus voting appears to be more stable. In all but the Fissure case study this method also reported similar performances to that obtained using the STAPLE and LSML algorithms.

By and large, when the variance between annotations is relatively low (for example in the Landslide case study in which the F$_1$-score differences range from \num{0.14} to \num{0.28}) the STAPLE and LSML methods provide GTs that report a performance within the middle of that reported by each of the individual annotations. Nevertheless, as noted above, this is not the case when annotation variance increases or few annotations are available (as in the Blood Vessel case study) and this seems to be in line with other studies \cite{Langerak10}. The SIMPLE algorithm was proposed to overcome these limitations when annotator uncertainty varies considerably \cite{Langerak10} and indeed, in these situations it does seem to offer an improvement (see, for example, the Segmentation and Blood Vessel case studies). Nevertheless, when the variance in annotator agreement is not so extreme, SIMPLE seems to result in an overestimation of performance (see the Fissure dataset for example).

The output of all of the detectors produced medium to high correlations with annotator agreement. It can be stated that a detector's performance increases as the agreement upon the segment increases and those detectors resulting in the lowest drop in correlation (from CCO to CCI) result in a lower $\bar{\textrm{P}}$-R curve spread. This seems intuitive as agreement should be higher for more obvious segments and, assuming that the detector is effective, these should also elicit the highest detector responses. This translates to increasingly higher $\bar{\textrm{P}}$-R curves as GTs with higher levels of agreement are used. Unexpectedly however, when the correlation between detector output and agreement increases from within segment locations (CCO) to the whole image (CCI), precision decreases in lower recall ranges. Surprisingly, this reduction in precision indicates an accurate detector---as agreement increases, lower-agreement segments are removed from the GT causing the detector to classify them as false positives. This could be an indication that some of the annotators have missed important segments, which the detector considers to be true positives, and providing these locations as feedback to the annotators for confirmation could be a way of improving GT reliability.

The image features included in this study account for a high proportion of the observed agreement (it should be kept in mind these features are not independent of each other), but capture only local, low-level information, ignoring any higher level and global queues and knowledge that the annotators exploit. Further evidence for this is provided by the agreement level GT curves, which generally show that there is a large difference between the detection rate of high and low agreement segments---detection of the lower is not a trivial matter and the decision most likely needs to be augmented with high-level information that is not exploited by these detectors.

In all but the Landslide case study it has been shown that the rank of a detector is dependent upon the GT used for evaluation. It can therefore be stated that the variance in performance observed when evaluating two detectors using different GTs is not equal and therefore, the relative difference in performance between detectors is dependent upon the GT used for evaluation. Three different rankings were observed in three of the four case studies. In one occasion the top ranked detector changed depending upon the GT, however, in most cases the top ranked detector remained constant. This is partly due to the fact that these top ranked detectors are considerably superior to the remaining and had their performance been closer this would not have been the case. The effects are most obvious in the Blood Vessel case study, in which the detector that produces the worst correlation with annotator agreement (MSF: $\text{CCO} = 0.3923$ and $\text{CCI} = 0.3573$) was placed second, third, and fourth in each of the three emergent rankings, even though it is clearly the worst performing of the evaluated detectors. Moreover, taking the \SI{50}{\percent} or \SI{75}{\percent} consensus GTs does not necessarily result in a detector ranking that is the consensus of the ranks obtained using the individual annotations (see, for example, Tables IVb and IVd). In fact, it can produce a ranking that has nothing in common with these individual rankings (Table IVb).

The largest minimum bound on error, $\bar{e}$, was found in the Blood Vessel case study although the Segmentation and Fissure case studies produced the lowest pairwise F$_1$ scores (in fact the agreement between the two annotators in the Blood Vessel case study is relatively high). This uncovers two peculiarities with Smyth's calculation (see Equation \eqref{eq:smyth}) when used with only two, and an odd number of, annotators: the maximum of $\bar{e}$ is reached when the maximum disagreement amongst the annotators takes place. On either side of this maximum $\bar{e}$ decreases symmetrically. First, when only two annotators are present, $N=2$, any disagreement results in the maximum of the function since $[N-\max\{A(x,y), N-A(x,y)\}]/N \in \{0, 0.5\}$. Secondly, when an odd number of annotators are present this term can not reach the theoretical maximum of \num{0.5}, and therefore all disagreements contribute less than in the case of two annotators. Thus although the F$_1$ score attests to greater agreement in the Blood Vessel case study, it receives a higher minimum bound on the error.

Finally, as has been shown in the Segmentation case study, the evaluation framework adopted in this domain, through accounting for variances observed in the annotations, yields a very optimistic estimate of  algorithm performance when compared to the traditional precision-recall evaluation framework.


\subsection{Recommendations}

Comparing annotators and deciding upon outliers based solely upon inter-annotator performance is not a reliable method even though it offers reasonable modelling of---what could be described as---the average performance when correctly implemented (the SIMPLE, and to some extent the LSML, algorithms for example). Several counter examples can be easily proposed, such as a situation in which all but one annotator is inaccurate, a case in which the accurate annotator would be deemed an outlier and removed. Furthermore, an inaccurate annotation could in fact contain all of the true positive positions, but have low specificity, other annotations may have low sensitivity and therefore removing the `outlier' implies discarding valuable information that may not be possible to infer using other means. As \citet{Smyth96} states ``without knowing GT one can not make any statements about the errors of an individual labeller''.

Overly simplistic methods to utilise all of the available annotations (voting) have been shown to fail. More sensitive algorithms, such as STAPLE, take a step in the right direction. Nevertheless, these algorithms still assume that the gold-standard ground truth can be inferred by measuring the performance of annotators in relation to each other. The most promising advances have started to integrate information derived from the image into the process, and it has been shown herein that these properties do correlate with annotator agreement. Care should be taken, however, as this produces a somewhat circulatory solution in which the image features used by the detection algorithms are also used to decide upon which segments the algorithms are evaluated. Furthermore, in some domains correlation strengths between annotator agreement and image features decrease when moving from within segment locations to the whole image. Demonstrating that these properties are not uniquely tied to the segments of interest and employing this source of information risks introducing false positive locations to the inferred GT.

In other fields of science, progress has been made on improving the rating of annotator performance by gathering meta-data along with the annotations. The Cooke method \cite{Cooke91} prescribes that the annotators are asked to estimate a interval of probable values along with their concrete answer, and furthermore they are also asked to answer multiple questions on topics from their field that have known answers. This information is used to weight the annotator's contribution in relation to their accuracy in this estimation and thus, has been shown to be more accurate than consensus voting \cite{Aspinall10}.

It is clear that evaluating upon different GTs, whether these are annotations or some merging thereof, reveals different trends in the performance of classification algorithms. Synonymously, different images reveal different algorithm strengths during evaluation and, as such, large datasets are used to smooth the differences and reveal the best overall performing algorithm. However laborious it may be, the presented work implies that an algorithm should also be evaluated using different GTs. While the presented study does not offer an ultimate solution for how those GTs should be combined the described analysis framework provides a means to quantify the spread of measured performance and test whether the observed differences in performance are significant or not.

The variance of the annotations, and thus the variance of the algorithm's measured performance, is indicative of the number of annotations that should be collected for accurate evaluation. The Landslide case study, for example, exhibits low annotator variance and this is reflected in the spread of $\bar{\textrm{P}}$-R curves, which are relatively tightly clustered. Performance bounds can therefore be reliably estimated with few annotations. The Segmentation annotations, in contrast, exhibit large variance, as do the resulting $\bar{\textrm{P}}$-R curves. Under these conditions (and those in which few annotations are available, such as in the Blood Vessel case study) it may not be possible to state with certainty whether one algorithm outperforms another and further studies with more annotations should be conducted.

Considering that in all of the evaluated datasets the Any-GT and high agreement level GTs (0.5-GT or 0.75-GT) appear to model the lower and upper bounds (respectively) on the spread of measured performance, this may offer a means of measuring the performance overlap between two algorithms, which would be characteristic of the confidence that can be attributed to any measured differences in performance. 

This approach accepts that there exists imperfections in the individual annotations, which are included in the Any-GT, but assuming that a perfect detector is created, these imperfections cause the performance to degrade and simply decreases the lower bound on performance (and therefore represents the uncertainty inherent in the problem). Furthermore, there is a high likelihood that these imperfections are removed at high agreement levels (since they are variations of individual annotators). The upper bound, therefore is stable with respect to these and the true, unknown, detector performance is contained somewhere within these bounds.

Finally, to be able to use such an approach, and to understand annotator variance within standard evaluation datasets, it should be made possible to determine which annotations each annotator produced, and to ensure a sufficient coverage of the dataset by the same annotators.


\section{Conclusions}
\label{sec:conclusions}

This paper set out to quantify the effects of obtaining ground truth data from multiple annotators in a computer vision setting. It has also taken some steps towards identifying which properties of the image are related to agreement amongst the annotators. Statistical analyses of the GTs in each case study lead to the quantification of the differences between the annotations. A number of gold-standard GT estimation methods were evaluated, including removing outlier annotations, and it was found that the STAPLE and LSML algorithms find a balance between all annotations when their variance is low. Ground truths formed by taking segments that any of the annotators marked and thresholding at \SI{50}{\percent} and \SI{75}{\percent} agreement, tend to form lower and upper bounds on detector performance. Performance measured when using the GT derived by removing outlier annotations and then taking the consensus vote approaches that of STAPLE and LSML in all but one of the case studies. It does, however, appear to be more stable when the annotations have high variance.

It can be concluded that the rank of a detector is highly dependent upon which GT estimation algorithm is used. In some cases the GTs calculated by voting resulted in a detector rank that is in discordance with each of the individual annotations. The $\bar{\textrm{P}}$-R curves obtained using the voted GTs also appear to be outliers when compared to those of the remaining GTs, suggesting that these commonly employed GT estimation methods overemphasise detector performance when compared to individual annotator opinion. Furthermore, under some conditions a detector whose output is poorly correlated with annotator agreement can be placed above those that have vastly better correlated outputs.

Therefore in addition to evaluating an algorithm over a data set that contains multiple images, it is concluded that an algorithm should also be evaluated using multiple ground truths. The variance of performance that is observed using these different ground truths can then be used to quantify the confidence in the differences between detectors. In situations in which there are few annotations available, or when the inter-annotator variance is high, further study into the nature of the problem should be conducted as these conditions imply that it is not possible to state that one algorithm outperforms another with any confidence. Therefore, whenever possible the intrinsic uncertainties of annotator judgements should be assessed before the evaluation of detection algorithms, since measures of absolute performance and relative ranking of detectors may vary considerably according to the GT employed.

The possibility of estimating a detector's true performance through the variability of annotator opinion would be an interesting avenue to follow. Assuming that performances derived using different GTs are observations of a hidden variable, it may be possible to estimate its true value---the gold standard performance. Much research is dedicated to inferring the gold-standard GT, however, this is a complex problem in which many assumptions need to be made, and the proposed approach may avoid some of these.

An additional question that is raised by this study is: which metric should be used to evaluate an estimated gold standard? Generally speaking the gold standard is unknown and therefore comparison is impossible. Restricting evaluation to individual annotations assumes high specificity and sensitivity. Removing annotations, however, assumes inability compared to the consensus, but do those removed represent true insight into the problem? One thing is clear, detector performance should not be used to evaluate an estimated gold-standard ground truth.


 \section*{Acknowledgement}
The participating annotators from LIVE, IPGS, and ICube (University of Strasbourg), and ITC (University of Twente) are gratefully acknowledged.



\end{document}